\PassOptionsToPackage{table,dvipsnames}{xcolor}
\documentclass[10pt,twocolumn,letterpaper]{article}

\usepackage{wacv}              

\usepackage{graphicx}
\usepackage{amsmath}
\usepackage{amssymb}
\usepackage{booktabs}
\usepackage{xcolor}
\usepackage{pifont}
\usepackage{booktabs}
\usepackage{multirow}

\usepackage{tikz}
\usepackage{booktabs}
\usepackage{amssymb}
\usepackage{xcolor}
\usepackage{subcaption}

\newcommand{\cmark}{\textcolor{green!70!black}{\ding{51}}} 
\newcommand{\xmark}{\textcolor{red!80!black}{\ding{55}}}   
\newcommand{\prismcafo}{PRISM-CAFO}

\newcommand{\second}[1]{\cellcolor{green!25}{#1}}   
\newcommand{\best}[1]{\cellcolor{green!60}{#1}} 

\usepackage[pagebackref,breaklinks,colorlinks]{hyperref}

\usepackage[capitalize]{cleveref}
\crefname{section}{Sec.}{Secs.}
\Crefname{section}{Section}{Sections}
\Crefname{table}{Table}{Tables}
\crefname{table}{Tab.}{Tabs.}

\usepackage{enumitem} 
\setlist[itemize]{leftmargin=12pt,labelsep=5pt,noitemsep,topsep=5pt}
\setlist[enumerate]{leftmargin=20pt,labelsep=5pt,noitemsep,topsep=5pt}
\newlist{inline}{enumerate*}{1}
\setlist[inline]{before=\unskip{: }, itemjoin={{; }}, itemjoin*={{; and }}, label={(\roman*)}}

\def\wacvPaperID{3274} 
\def\confName{WACV}
\def\confYear{2025}

\begin{document}

\title{	
PRISM-CAFO: Prior-conditioned Remote-sensing Infrastructure Segmentation and Mapping for CAFOs}


\author{
Oishee Bintey Hoque\textsuperscript{\rm 1,\rm *}, Nibir Chandra Mandal\textsuperscript{\rm 1,\rm *}, Kyle Luong$^{1}$, Amanda Wilson$^2$, Samarth Swarup$^2$,\\ Madhav Marathe$^{1,2}$, Abhijin Adiga$^{2}$ \\
$^1$Dept. of Computer Science, University of Virginia \\
$^2$Biocomplexity Institute, University of Virginia}

\maketitle

\begin{abstract}
Large-scale livestock operations pose significant risks to human health and the environment, while also being vulnerable to threats such as infectious diseases and extreme weather events. As the number of such operations continues to grow, accurate and scalable mapping has become increasingly important. In this work, we present an infrastructure-first, explainable pipeline for identifying and characterizing Concentrated Animal Feeding Operations (CAFOs) from aerial and satellite imagery. Our method ($i$) detects candidate infrastructure (e.g., barns, feedlots, manure lagoons, silos) with a domain-tuned YOLOv8 detector, then derives SAM2 masks from these boxes and filters component-specific criteria; ($ii$) extracts structured descriptors (e.g., counts, areas, orientations, and spatial relations) and fuses them with deep visual features using a lightweight spatial cross-attention classifier; and ($iii$) outputs both CAFO type predictions and mask-level attributions that link decisions to visible infrastructure. Through comprehensive evaluation, we show that our approach achieves state-of-the-art performance, with Swin-B+PRISM-CAFO surpassing the best performing baseline by up to \textbf{15\%}. Beyond strong predictive performance across diverse U.S. regions, we run systematic gradient–activation analyses that quantify the impact of domain priors and show how specific infrastructure (e.g., barns, lagoons) shapes classification decisions. We release code, infrastructure masks, and descriptors to support transparent, scalable monitoring of livestock infrastructure, enabling risk modeling, change detection, and targeted regulatory action.
\textbf{Github: } \href{https://github.com/Nibir088/PRISM-CAFO}{https://github.com/Nibir088/PRISM-CAFO}.

\end{abstract}

\vspace{-.5cm}
\section{Introduction}
\label{sec:intro}
\begin{figure}
    \centering
    \includegraphics[width=\linewidth]{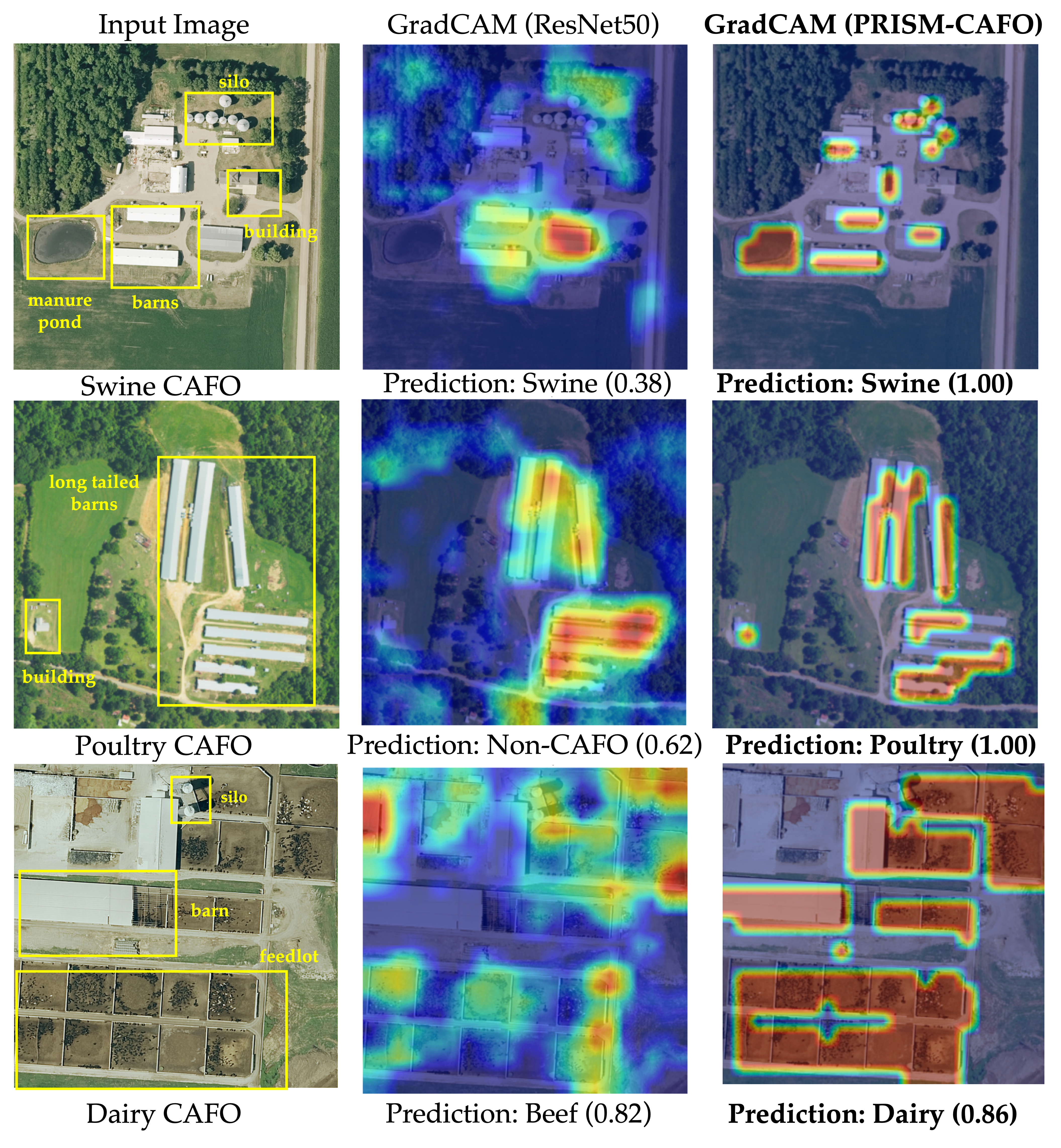}
    \caption{\prismcafo{}'s candidate-infrastructure-focused approach not only leads to better identification
    of CAFOs, but also captures key infrastructure in the process with higher confidence scores (based on GradCAM~\cite{selvaraju2017grad}).
    }
    \label{fig:placeholder}
\end{figure}
Concentrated Animal Feeding Operations (CAFOs) are large industrial facilities that confine thousands of animals in small areas~\cite{moses2017industrial,9832662}. They are vulnerable to infectious diseases that can spill over to humans and generate substantial waste, contributing to nutrient runoff, degraded water quality, and greenhouse gas emissions~\cite{moses2017industrial}. The ongoing Highly Pathogenic Avian Influenza (HPAI) epidemic highlights the risks associated with high-density livestock operations~\cite{prosser2024using,humphreys2020waterfowl,adiga2024high,nguyen2025emergence}, making CAFOs a critical One Health concern. Despite accounting for over half of US livestock production, their locations and prevalence remain difficult to track due to regulatory gaps and limited transparency~\cite{GurianSherman2008,Hribar2010, HandanNader2021}. Mapping CAFO locations and attributes--facility size, livestock type, and key components such as barns and manure ponds--is essential for environmental risk assessment and epidemiological modeling.
Remote sensing and deep learning are increasingly used to map agricultural infrastructure. However, current CAFO-identification methods have notable limitations: ($i$) patch-based CNNs rely on small, localized datasets and often detect only a few livestock types~\cite{9832662,handan2019deep}; and ($ii$) indirect models using methane, land use, or socio-environmental variables~\cite{zhu2022meter,saha2025machine} do not provide facility-level characterization. However, recent public releases of state-level CAFO data (including livestock type, location, herd size, and infrastructure details) create a timely opportunity for the AI community to develop more comprehensive, data-driven models for large-scale CAFO detection and analysis.

\begin{figure}[t]
  \centering
  \begin{subfigure}{1\linewidth}
    \includegraphics[width=\linewidth]{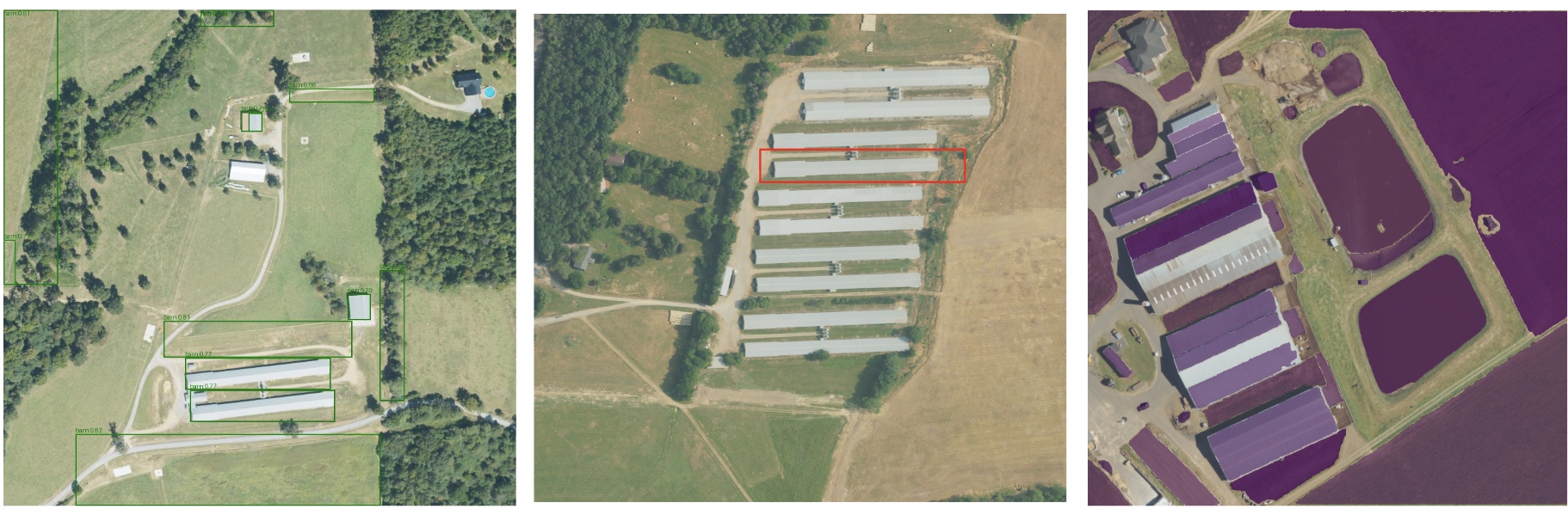}
    \caption{Illustration of noisy and incomplete bounding boxes from open-vocabulary models: ($i$) Grounding DINOv2 (left) and ($ii$) Florence-2 (center), and class-agnostic masks from SAM2 (right).}
    \label{fig:noise}
  \end{subfigure}\hfill
  \begin{subfigure}{1\linewidth}
    \includegraphics[width=\linewidth]{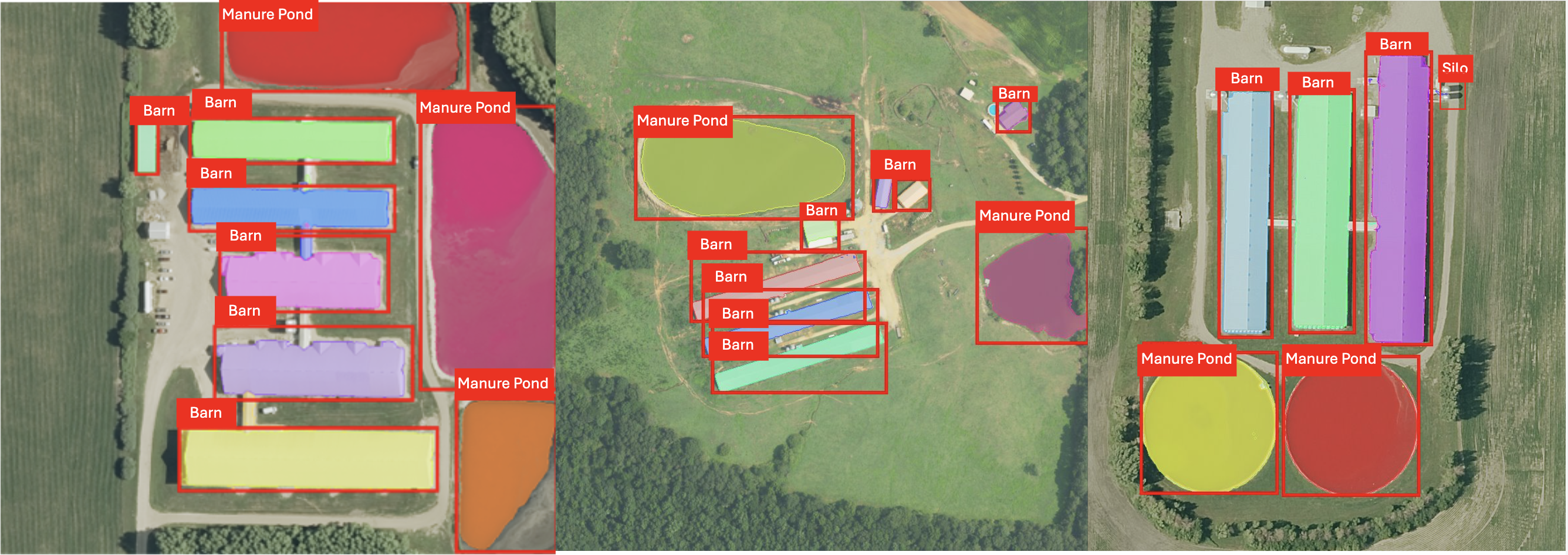}
    \caption{Illustration of outputs from the infrastructure candidate generation module, showing accurate proposals with strong feature representations.}
    \label{fig:yolo}
  \end{subfigure}\hfill
  \caption{Qualitative comparison of existing object identification approaches with our domain-specific module.}
  \label{fig:example}
\end{figure}

\paragraph{Challenges.}  
Livestock operations include elements such as barns, silos, and manure ponds, and their size, composition, and layout can vary widely—even within the same livestock type. Many components, like barns, also appear in non-CAFO settings (e.g., warehouses), making visual identification difficult. Although accurately inferring livestock type requires understanding a farm’s specific infrastructure composition, labeled data is limited relative to this complexity. Survey data from sources like the Census of Agriculture~\cite{agcensus2022}, while useful for locating facilities, provides no component-level annotations, leaving a critical gap for fine-grained analysis. Advanced class-agnostic segmentation models, such as SAM2~\cite{kirillov2023sam}, can generate high-quality object masks but provide no semantic labels, requiring domain knowledge to filter relevant structures by their shapes, sizes, and positions. In contrast, open-vocabulary methods (e.g., OVSeg~\cite{liang2023ovseg}, OWL-ViT~\cite{minderer2023owlv2}) can assign categories from text prompts, but they struggle in specialized domains where distinguishing CAFO-specific infrastructure (e.g., barns, manure lagoons) from visually similar non-CAFO structures (e.g., warehouses, retention ponds) demands context-aware expertise (See Fig.~\ref{fig:noise}). These challenges underscore the need for domain-adapted datasets and methods that integrate expert knowledge and supporting statistics to achieve reliable CAFO detection and characterization. 

\paragraph{Contributions.}

\begin{figure*}
    \centering
    \includegraphics[width=\linewidth]{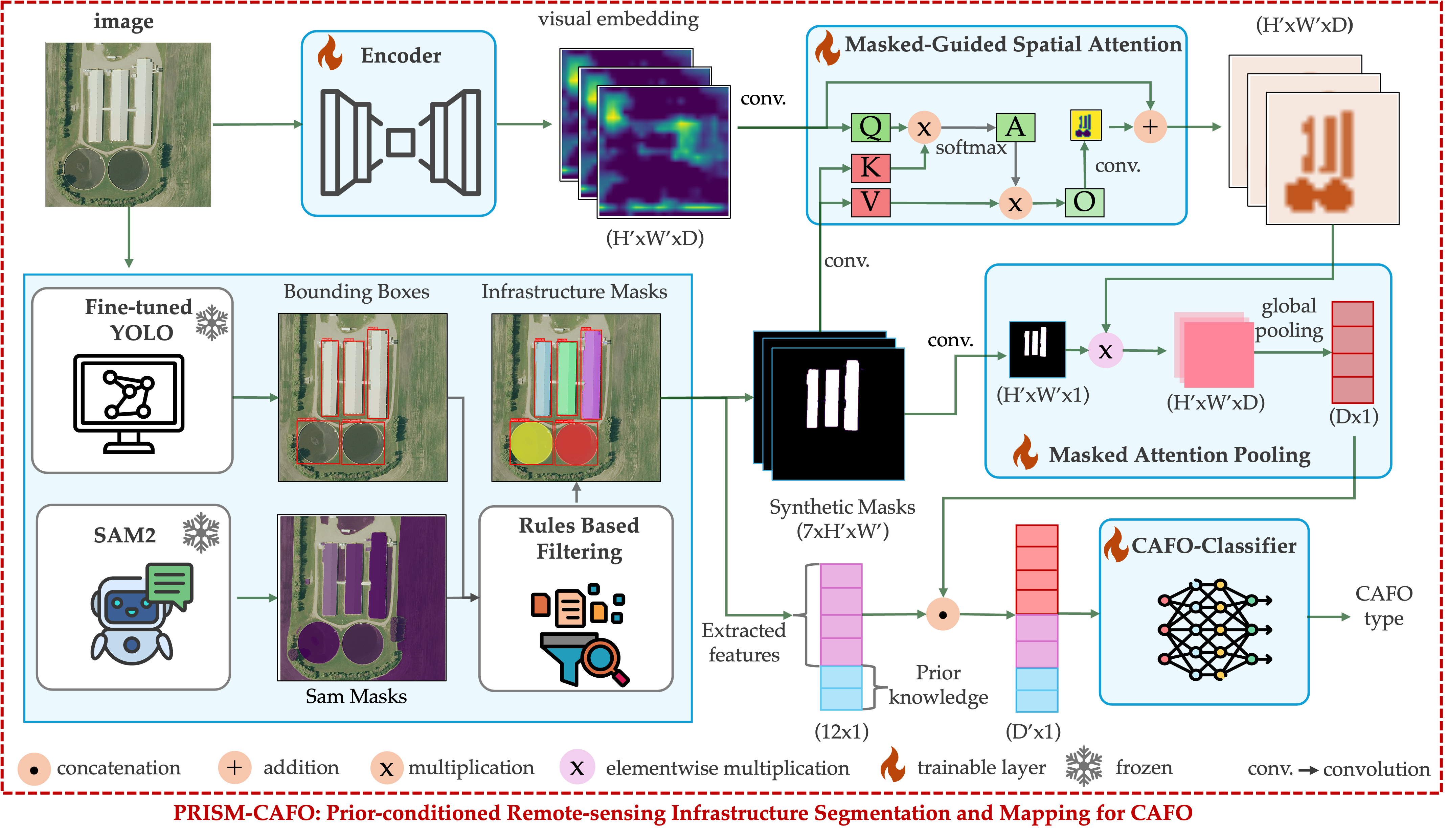}
    \caption{The schematic of the \prismcafo{} framework. Refined YOLO and SAM2 masks form synthetic priors that modulate visual embeddings via masked-guided spatial attention, producing prior-aware features for accurate CAFO classification. }
    \label{fig:pipeline}
\end{figure*}

\begin{itemize}
    \item We bootstrap a domain-tuned YOLOv8 with manually verified boxes and fuse its proposals with SAM2 masks and geometric filtering (See Fig.~\ref{fig:pipeline}), yielding well-verified, semantically consistent infrastructure candidates for geometry- and relation-aware CAFO analysis (See Fig.~\ref{fig:yolo}).
    

    \item We develop a \textbf{mask-guided spatial attention} classifier that fuses infrastructure-grounded features with engineered domain cues (See Fig.~\ref{fig:pipeline}), enabling geometry- and relation-aware CAFO categorization. We further propose \textbf{an attention-weighted pooling mechanism} that aggregates features only over infrastructure regions, producing interpretable, layout-sensitive embeddings that outperform global pooling. Unlike vision-only models, it achieves competitive accuracy while providing object-level, spatially grounded interpretability.
    
    
    \item We apply our framework to multiple image classification models. Our approach delivers the best performance among all evaluated baselines, with Swin-B+\prismcafo{} improving \textbf{F1-score by up to 15\%}.

     \item Our framework supports mask-level attributions and Grad-CAM-style visual explanations, linking model decisions directly to specific barns, lagoons, and feedlots. In addition, we provide a systematic analysis of feature and infrastructure-mask importance (through gradient-activation~\cite{wang2021feature}).
     
\end{itemize}
Component-level CAFO characterization unlocks applications beyond classification, including supply-chain modeling, disease surveillance, and agricultural intensification analysis. By converting raw pixels into a queryable object-level database, we provide a valuable resource for researchers and policymakers.




\section{Related Work}
\paragraph{Mapping CAFOs.} 
Mapping Concentrated Animal Feeding Operations~(CAFOs) is a relatively
recent and evolving research area. Robinson et
al.~\cite{9832662} train convolutional neural network (CNN) models on USDA
NAIP 1-meter aerial imagery, using labeled CAFO locations from four states
for training and validation. Handan et al.~\cite{handan2019deep} apply
CNNs to identify swine and poultry operations in North Carolina.
METER-ML~\cite{zhu2022meter} addresses a related but distinct task mapping
methane-related facilities using multi-resolution imagery~(NAIP and
Sentinel-2) across a broader geographic scope. Chugg et al.~\cite{CHUGG2021102463} track CAFO expansion using building segmentation and change detection on PlanetScope imagery. In contrast, Saha et al.~\cite{saha2025machine} use a random forest on socio-environmental features across 18 states, without satellite data.


\noindent\textbf{Object-centric scene recognition.}
Scene recognition studies show that scenes can be understood from their constituent objects. Heikel and Espinosa-Leal~\cite{heikel2022tfidf} treat detected objects as TF–IDF tokens, while Song et al.~\cite{song2024interobject} model discriminative inter-object relations with graphs. Kapoor et al.~\cite{exploiting2024objseg} combine detections with segmentation cues to leverage layout and part boundaries, and Zhou et al.~\cite{zhou2025hierarchical} integrate hierarchical objectness and spatial structure to improve scene categorization.

\noindent\textbf{Scene graph generation and relation reasoning.}
While Scene Graph Generation (SGG) detects objects and pairwise relations as mid-level scene representation, application is limited. Recent methods~\cite{li2024ovsgg,li2024predicate} focus on relationship prediction. However, for livestock operations, these relations are often ambiguous and, critically, no labeled data exists to define them.




\noindent\textbf{Localizing Objects.}
Promptable, class-agnostic segmenters like SAM2 produce high-quality masks but lack categories and exhaustive localization~\cite{ravi2024sam2,kirillov2023sam}. Open-vocabulary models~\cite{adiga2024high,minderer2023owlv2,liang2023ovseg,zou2023segment} broaden label space via text prompts but remain brittle for domain-specific taxonomies, as noted in medical and aerial imagery~\cite{cheng2023sammed,ke2024universal}. General-purpose encoders (CLIP, DINOv2, Florence-2)~\cite{radford2021clip,oquab2023dinov2,xiao2024florence2,yuan2021florence} offer transferable features but are not tuned to close-set categories like CAFO. Detector-anchored pipelines are more reliable: YOLOv8~\cite{jocher2023yolov8} provides efficient, domain-tunable proposals, and weakly supervised specialization~\cite{ding2024specializeddet} can adapt detectors to novel classes. We therefore use a hybrid approach: a domain-tuned YOLOv8 generates proposals and labels, SAM2 refines instance masks, and human-in-the-loop pseudo-labeling/self-training~\cite{lee2013pseudolabel,xie2020noisystudent} scales a small set of verified boxes into over $130$K high-quality masks.

\section{Methodology}
We present a detection–segmentation–classification framework that fuses \emph{infrastructure-grounded vision} with \emph{symbolic and geo-statistical priors} for CAFO type prediction. The pipeline (Fig.~\ref{fig:pipeline}) comprises: (i) \textbf{Synthetic Infrastructure Mask Generation}, (ii) \textbf{Feature Engineering with Domain Priors}, and (iii) \textbf{Attention-Based Classifier}.

\subsection{Synthetic Infrastructure Mask Generation}
\label{sec:mask}
Accurate CAFO categorization depends on the presence and spatial organization of infrastructure (e.g. barns, lagoons, silos, feedlots, etc). Infrastructure-level masks explicitly ground the model in these structures, enabling geometry- and relation-aware reasoning.

\noindent\textbf{Infrastructure Detection (Region Proposals).}
Given an image $I\!\in\!\mathbb{R}^{H\times W\times C}$, a fine-tuned YOLOv8~\cite{jocher2023yolov8} detector $g$ generates proposals:
\[
\{(o_i,\,\,k_i)\}_{i=1}^{N} = g(I),
\]
where $o_i$ is a bounding box and $k_i\!\in\!\{1,\dots,K\}$ the infrastructure category ($K{=}7$). Detection provides a discrete, structured hypothesis set for downstream segmentation and mask fusion, which is crucial because CAFO labels depend on co-occurrence of specific structures. 

To fine-tune the YOLO model, we generate a reliable training set: (i) we first selected 114 CAFO sites and applied the SAM2 model to produce candidate masks ($\sim 5000$ masks were generated); and ($ii$) bounding boxes were drawn around each mask, and every box was manually labeled. In total, 1069 boxes corresponded to true infrastructure components (barns, ponds, silos, etc.), while the remaining boxes $\sim4000$ masks were labeled as negatives. This curated set of positive and negative examples was then used to fine-tune the YOLOv8 detector.

\noindent\textbf{Infrastructure Segmentation.}\label{sec:mask} For each detection box \(o_i\), we apply SAM2\cite{ravi2024sam2} to generate a binary mask \(m_i \in \{0,1\}^{H\times W}\) conditioned on \(o_i\), ensuring the mask is spatially anchored to the detected region while capturing fine-grained object boundaries. Segmentation provides the spatial precision (boundaries/shape) required for area, compactness, and proximity computations.

\noindent\textbf{Geometric Filtering.}
We evaluate each candidate \((m_i, o_i, k_i)\)—a mask, detection box, and component type—using the following geometric functionals:
\begin{itemize}
    \item \text{Containment ($\phi$):} measures how much of the mask lies within the detection box.
    \item Coverage ($\psi$): measures how much of the detection is explained by the mask.
    \item Rectangularity ($\rho$): measures how compactly the mask fills its bounding rectangle.
    \item Relative Size ($\sigma$): measures the scale of the mask relative to the detection box.
\end{itemize}
\[
\phi(m, o) = \frac{A(m \cap o)}{A(m)};\qquad \psi(m, o) = \frac{A(m \cap o)}{A(o)} 
\]
\[
\rho(m) = \frac{A(m)}{A(\mathrm{bbox}(m))};\qquad \sigma(m, o) = \frac{A(\mathrm{bbox}(m))}{A(o)}
\]







These are used to filter component candidates based on expected shape and spatial consistency:
\textit{(i)} Barns are expected to be well-contained, rectangular, and tightly bounded, i.e., \( \phi(m,o) \ge \tau_{\text{in}}^{\text{barn}} \) and \( \rho(m) \ge \tau_{\text{rect}}^{\text{barn}} \), with \( \tau_{\min}^{\text{barn}} \le \sigma(m,o) \le \tau_{\max}^{\text{barn}} \). \textit{(ii)} Ponds are amorphous and should maximally cover the detection area; we retain only the candidate with maximal coverage \( m^\star = \arg\max_m \psi(m,o) \) and require \( \psi(m^\star,o) \ge \tau_{\text{cover}}^{\text{pond}} \). \textit{(iii)} Silos are small, compact objects expected to lie within the detection box, enforcing \( \phi(m,o) \ge \tau_{\text{in}}^{\text{silo}} \), \( \rho(m) \ge \tau_{\text{rect}}^{\text{silo}} \), and \( \sigma(m,o) \le \tau_{\max}^{\text{silo}} \). \textit{(iv)} Feedlots are large open enclosures and must overlap sufficiently, with \( \psi(m,o) \ge \tau_{\text{ov}}^{\text{feed}} \) and \( \sigma(m,o) \ge \tau_{\min}^{\text{feed}} \). \textit{(v)} Silage/Storage requires minimal consistency between the mask and the detection region, \( \psi(m,o) \ge \tau_{\text{ov}}^{\text{silage}} \). \textit{(vi)} All other categories apply a general overlap rule, \( \psi(m,o) \ge \tau_{\text{ov}}^{\text{default}} \).

\noindent Note that these thresholds \(\{\tau\}\) were calibrated by manually inspecting the alignment between SAM2-generated masks and YOLO detections, ensuring geometric plausibility grounded in real-world CAFO infrastructure. 

\noindent\textbf{}{Synthetic Masks.}
Filtered binary masks are combined across proposals to produce  multi-channel masks
\[
C \in \{0,1\}^{H \times W \times K}, \qquad
C(:,:,k) = \max_{i:\,k_i=k} m_i,
\]
where each channel $M(:,:,k)$ encodes the spatial footprint of infrastructure type $k$. These masks provide explicit localization of barns, ponds, silos, feedlots, and related structures, serving as structured inputs that guide spatial attention and enable geometry and relation-aware feature extraction in downstream modeling. 

\subsection{Feature Engineering with Domain Priors}
Beyond learned appearance, explicit descriptors capture infrastructure composition, spatial relations, and regional contexts that are determinative for CAFO types. We construct a 12 dimensional feature vector.

\noindent\textbf{Infrastructure Areas (7 features).}
Let $C_k=C(:,:,k)$. Define a differentiable soft area,
\[
\operatorname{area}(C_k)=\tfrac{1}{HW}\!\sum_{x,y} C_k(x,y), \qquad
A_{\mathrm{all}}=\sum_{j=1}^{K}\operatorname{area}(C_j)+\varepsilon .
\]
We use per-infrastructure area ratios,
\[
r_k=\frac{\operatorname{area}(C_k)}{A_{\mathrm{all}}}, \quad k\in{1,\dots,7},
\]
which summarize functional footprints and enable geometry-aware priors (e.g., barn-to-pond share).

\noindent\textbf{Barn--Pond Proximity (1 feature).} We measure barn--pond adjacency via the symmetric Chamfer Distance~\cite{bakshi2023near} between \textit{\(C_{\text{barn}}\) and \(C_{\text{pond}}\), i.e., \(d_{\mathrm{bp}}=\operatorname{Chamfer}(C_{\text{barn}},\,C_{\text{pond}})\in\mathbb{R}_{\ge 0}\)}, explicitly encoding a key structural indicator (e.g., barn--lagoon collocation for swine).
\\
\noindent\textbf{County-Level Class Priors (4 features).}
From the Census of Agriculture~\cite{agcensus2022} we obtain normalized fraction of livestock operations corresponding to each livestock type 
at the county level:
\(
\mathbf{q}=[q_{\text{swine}},\,q_{\text{poultry}},\,q_{\text{dairy}},\,q_{\text{beef}}]\in\Delta^3,
\)
For each image patch, the values of the corresponding county are 
appended to the feature vector.
\\
\noindent\textbf{Prior Feature Vector (12-D).} We concatenate composition, relation, and geo-statistical cues, forming \(\mathbf{f} = [\,r_1,\ldots,r_7,\, d_{\mathrm{bp}},\, \mathbf{q}\,] \in \mathbb{R}^{12}\), then standardize each feature (training-set mean/variance) before fusion with learned embeddings.
\subsection{Deep Attention-Based Classifier}
Engineered priors are complementary to learned representations that capture texture, context, and higher-order patterns. We fuse \emph{mask-guided spatial attention} with \emph{mask-driven attention pooling} to produce an infrastructure-grounded embedding.
\\
\noindent\textbf{Visual Encoder.}
A CNN/Transformer backbone $f_{\theta}$ maps the image to a feature tensor:
\[
E=f_{\theta}(I)\in\mathbb{R}^{H'\times W'\times D}.
\]
where $\theta$ denotes the learnable parameters of the encoder. Here, $H'$ and $W'$ are the spatial dimensions of the encoded feature map.  This provides high-level semantics under viewpoint, seasonality, and sensor variations.
\\
\noindent\textbf{Mask-Guided Spatial Attention.}
We integrate mask information into the feature encoding via a spatial attention mechanism. Given visual features $E \in \mathbb{R}^{ H'\times W'\times D}$ and infrastructure masks $C \in \mathbb{R}^{H\times W\times K}$, we first resize $C$ to match the spatial resolution of $E$, yielding $C' \in \mathbb{R}^{H'\times W'\times K}$. The module then projects features and masks into query, key, and value representations:
\[
Q = W_q \circledast E,\quad K = W_k \circledast C',\quad V = W_v \circledast C',
\]
where $\circledast$ denotes $1{\times}1$ convolutions. Here, $W_q, W_k, W_v$ are learnable parameters, $Q \in \mathbb{R}^{H'\times W'\times d_a}$, $K \in \mathbb{R}^{H'\times W'\times d_a}$, and $V \in \mathbb{R}^{H'\times W'\times D}$. Here $d_a$ denotes the attention dimension.

Flattening the spatial dimensions yields $Q \in \mathbb{R}^{H'W'\times d_a}$, $K \in \mathbb{R}^{H'W'\times d_a}$, and $V \in \mathbb{R}^{H'W'\times D}$. Attention scores are then computed as
\[
A = \operatorname{softmax}\!\left(\frac{Q K^ \top}{\sqrt{d_a}}\right) \in \mathbb{R}^{H'W'\times H'W'}.
\]
The attended output is $O = A V \in \mathbb{R}^{ H'W'\times D},$ which is reshaped back to $\mathbb{R}^{H'\times W'\times D}$.

Finally, a bottleneck projection $W_o$ (two $1{\times}1$ convolutions with a hidden Rectified Linear Unit (ReLU) layer) maps $O$ back into the visual feature space. The residual-enhanced fusion is,
\[
E' = W_o(O) + E,
\]
which preserves the original visual semantics while injecting mask-informed attention. This module ensures that the encoder emphasizes infrastructure-consistent regions while remaining lightweight and computationally efficient.
\\
\noindent\textbf{Mask Attention Pooling.}
Given fused features $E'$ and reshaped infrastructure masks $C'$, an attention subnetwork--comprising a $3{\times}3$ convolution, ReLU activation, and another $3{\times}3$ convolution followed by sigmoid--produces a scalar attention map:
\[
A = \sigma\!\big(W_2 \circledast \mathrm{ReLU}(W_1 \circledast C')\big) \;\in\; [0,1]^{H'\times W'},
\]
where $\circledast$ denotes convolution and $C'$ denotes the resized mask. This map highlights regions aligned with infrastructure evidence. We then compute attention-weighted global average pooling:
\[
\mathbf{z}=\frac{\sum_{x,y} A(x,y)\,E'(x,y)}{\sum_{x,y} A(x,y)+\varepsilon} \;\in\; \mathbb{R}^{D}.
\]
By emphasizing discriminative structures (e.g., barn–lagoon clusters) while suppressing background, this pooling strategy yields more robust representations than uniform averaging.
\\
\noindent\textbf{CAFO Classifier.}
We concatenate learned and engineered modalities so that implicit appearance cues and explicit priors jointly determine the CAFO type.
\[
\mathbf{h}=\big[\mathbf{z}\,\|\,\mathbf{f}\big],\qquad
\hat{\mathbf{y}}=W\mathbf{h}+\mathbf{b},\qquad
\hat{\mathbf{p}}=\operatorname{softmax}(\hat{\mathbf{y}}),
\]
where, $\hat{\mathbf{p}}$ is the predicted probability.

\begin{table*}[ht]
\centering
\small
\setlength{\tabcolsep}{6pt}
\begin{tabular}{lcccccccccc} 
\toprule
\multirow{2}{*}{\textbf{Model}} & \multicolumn{5}{c}{\textbf{Random (80\%-20\%) Split}} & \multicolumn{5}{c}{\textbf{Spatial Split}} \\
\cmidrule(lr){2-6} \cmidrule(lr){7-11}
 & Swine & Poultry & Dairy & Beef & Neg & Swine & Poultry & Dairy & Beef & Neg \\
\midrule
CLIP (ViT-B/32)             & 0.869 & 0.719 & 0.449 & 0.482 & 0.952 & 0.673 & 0.636 & 0.280 & 0.143 & 0.951 \\
CLIP+\prismcafo{}             & 0.854$\downarrow$ & 0.707$\downarrow$ & 0.418$\downarrow$ & 0.428$\downarrow$ & 0.962$\uparrow$ & 0.550$\downarrow$ & 0.554$\downarrow$ & 0.341$\uparrow$ & 0.150$\uparrow$ & 0.968$\uparrow$ \\
\midrule
DINOv2 ViT-B/16             & 0.900 & 0.793 & 0.577 & 0.624 & 0.963 & 0.758 & 0.737 & 0.406 & 0.182 & 0.958 \\
DINOv2+\prismcafo{}           & 0.899$\downarrow$ & 0.808$\uparrow$ & 0.579$\uparrow$ & 0.547$\downarrow$ & 0.981$\uparrow$ & 0.735$\downarrow$ & 0.747$\uparrow$ & 0.446$\uparrow$ & 0.159$\downarrow$ & 0.975$\uparrow$ \\
\midrule
RemoteCLIP (ViT-B/32)       & 0.875 & 0.755 & 0.466 & 0.489 & 0.960 & 0.700 & 0.708 & 0.354 & 0.188 & 0.958 \\
RemoteCLIP+\prismcafo{}       & 0.850$\downarrow$ & 0.673$\downarrow$ & 0.499$\uparrow$ & 0.353$\downarrow$ & 0.960$=$ & 0.603$\downarrow$ & 0.664$\downarrow$ & 0.311$\downarrow$ & 0.111$\downarrow$ & 0.971$\uparrow$ \\
\midrule
Swin-B                      & 0.881 & 0.772 & 0.573 & 0.556 & 0.958 & 0.692 & 0.706 & 0.446 & 0.159 & 0.951 \\
Swin-B+\prismcafo{}           & 0.928$\uparrow$ & 0.848$\uparrow$ & \best{0.678}$\uparrow$ & 0.689$\uparrow$ & 0.981$\uparrow$
                            & \best{0.874}$\uparrow$ & 0.758$\uparrow$ & \best{0.584}$\uparrow$ & 0.164$\uparrow$ & 0.986$\uparrow$ \\
\midrule
ViT-B/16                    & 0.895 & 0.775 & 0.578 & 0.596 & 0.962 & 0.769 & 0.730 & 0.439 & 0.249 & 0.955 \\
ViT-B/16+\prismcafo{}         & 0.917$\uparrow$ & 0.839$\uparrow$ & 0.612$\uparrow$ & 0.621$\uparrow$ & \best{0.989}$\uparrow$
                            & 0.715$\downarrow$ & 0.735$\uparrow$ & 0.313$\downarrow$ & 0.222$\downarrow$ & 0.979$\uparrow$ \\
\midrule\midrule
EfficientNet-B0             & 0.863 & 0.744 & 0.466 & 0.573 & 0.942 & 0.712 & 0.695 & 0.200 & 0.190 & 0.941 \\
EfficientNet-B0+\prismcafo{}  & \second{0.929}$\uparrow$ & \second{0.864}$\uparrow$ & 0.650$\uparrow$ & 0.680$\uparrow$ & 0.980$\uparrow$
                            & \second{0.855}$\uparrow$ & 0.800$\uparrow$ & \second{0.540}$\uparrow$ & \best{0.322}$\uparrow$ & 0.967$\uparrow$ \\
\midrule
EfficientNet-B3             & 0.859 & 0.722 & 0.519 & 0.548 & 0.936 & 0.642 & 0.737 & 0.350 & 0.205 & 0.936 \\
EfficientNet-B3+\prismcafo{}  & \best{0.939}$\uparrow$ & \best{0.872}$\uparrow$ & 0.629$\uparrow$ & \best{0.739}$\uparrow$ & 0.798$\uparrow$
                            & 0.780$\uparrow$ & 0.539$\uparrow$ & 0.539$\uparrow$ & \second{0.306} $\uparrow$ & 0.975$\uparrow$ \\
\midrule
ResNet18                    & 0.865 & 0.702 & 0.452 & 0.544 & 0.933 & 0.608 & 0.657 & 0.220 & 0.169 & 0.921 \\
ResNet18+\prismcafo{}         & 0.916$\uparrow$ & 0.829$\uparrow$ & \second{0.657}$\uparrow$ & \second{0.709}$\uparrow$ & \second{0.984}$\uparrow$
                            & 0.852$\uparrow$ & \best{0.868}$\uparrow$ & 0.522$\uparrow$ & 0.253$\uparrow$ & \best{0.991}$\uparrow$ \\
\midrule
ResNet50                    & 0.866 & 0.735 & 0.486 & 0.551 & 0.937 & 0.680 & 0.685 & 0.301 & 0.124 & 0.932 \\
ResNet50+\prismcafo{}         & 0.876$\uparrow$ & 0.811$\uparrow$ & 0.642$\uparrow$ & 0.684$\uparrow$ & 0.963$\uparrow$
                            & 0.833$\uparrow$ & \second{0.850}$\uparrow$ & 0.531$\uparrow$ & 0.090$\downarrow$ & \second{0.987}$\uparrow$ \\
\bottomrule
\end{tabular}
\caption{Per-class F1 scores for random (80–20) and spatial splits. Best values are highlighted in dark green and second-best in light green (per column). For each baseline, the \prismcafo{} variant is annotated with $\uparrow$ (improved), $\downarrow$ (worse), or $=$ (no change).}
\label{tab:overall_vs_fold2}
\end{table*}

\section{Experiments}

\subsection{Datasets}


\paragraph{Image Level Dataset.} Existing CAFO datasets are constrained by limited geographic scope, single-species focus, or coarse annotations that omit key infrastructure, restricting their applicability for scalable and fine-grained analysis~\cite{handan2019deep,ehrenpreis2021nmcafo,zhu2022meter,CHUGG2021102463,saha2025machine}. To overcome these limitations, we generated a strongly annotated benchmark for CAFO detection across $20$ U.S. states, comprising over 38{,}000 high-resolution image patches (833$\times$833 pixels, 0.6 m resolution) sourced from the USDA National Agriculture Imagery Program (NAIP)~\cite{USGS_NAIP}. Our dataset spans all regions included in prior resources such as METER-ML and RegLab,
but extends their coverage substantially by incorporating multiple livestock types (swine, poultry, dairy, beef) and providing fine-grained infrastructure annotations across a broader set of states. A
table of all data sources used is in the supplement. CAFO locations were compiled from multiple federal and state inventories and manually refined to align with visible infrastructure. Negative samples were generated using stratified land-cover masks to ensure realistic background variation. The dataset contains 1899 beef, 3615 poultry, 1167 dairy, 10894 swine, and 20771 negative samples. For evaluation, we adopt two complementary strategies: (i) a \emph{random 80--20 split}, with 80\% of samples for training/validation and 20\% for testing, and (ii) a \emph{spatial split} to assess out-of-distribution robustness, where 25\% of states (4 randomly chosen with $\geq$10 CAFOs) are held out entirely for testing. This dual design supports both standard in-distribution benchmarking and rigorous geographic generalization tests.

\noindent\textbf{Object-Level Dataset.} To our knowledge, no prior dataset provides object-level annotations of CAFO infrastructure, which are essential for fine-grained modeling and evaluation. To address this gap, we first manually verified and labeled 5{,}000 bounding boxes (bboxes) initialized from SAM2 masks, and used these verified bboxes to train an initial YOLOv8 detector. We then manually verified an additional 2{,}500 YOLOv8-predicted bboxes. Manual checks indicated high accuracy in infrastructure identification, label assignment, and structural delineation. After achieving strong detector performance, we retrained YOLOv8 with the verified data and ran inference over the full image set, generating predicted SAM2 masks using the methodology described in Section~\ref{sec:mask}. In total, this human-in-the-loop bootstrapping process produced approximately 130k\ high-quality instance masks.

\subsection{Comparison with Baselines}
We evaluate diverse architectures including CNN backbones (ResNet~\cite{he2016resnet}, EfficientNet~\cite{tan2019efficientnet}) and Transformer backbones (ViT~\cite{dosovitskiy2021vit}, Swin~\cite{liu2021swin}, DINOv2~\cite{oquab2023dinov2}, RemoteCLIP~\cite{liu2023remoteclip}) to represent efficient general-purpose models alongside domain-specific approaches tailored for remote sensing applications. This comprehensive selection enables direct comparison between standard computer-vision architectures and specialized satellite-imagery models. We report our approach alongside state-of-the-art baselines in Table~\ref{tab:overall_vs_fold2}.

\noindent\textbf{Random (80--20) Split.} 
For \emph{CNN backbones}, \prismcafo{} shows strong gains: Dairy improves by +39\%, Beef by +30\%, and Poultry by +18\% on ResNet and EfficientNet models. The best CNN (EfficientNet-B3+\prismcafo{}) achieves +9\% over its baseline. 
In contrast, for \emph{Transformer backbones}, DINOv2 ViT-B/16 performs better than our approach, but Swin-B+\prismcafo{} yields broader boosts, e.g., Dairy (+9\%) and Beef (+24\%). RemoteCLIP, despite being tuned for remote sensing, lags behind other ViTs but still benefits from priors. Overall, the best baseline (DINOv2) is surpassed by EfficientNet-B3+\prismcafo{} with a +4\% gain.  

\noindent\textbf{Spatial Split.} 
For \emph{CNN backbones}, gains are largest under distribution shift: Dairy improves by +170\%, Beef by +70\%, and Poultry by +32\%, underscoring the role of priors in hard classes. Similarly, for \emph{Transformer backbones}, Swin-B+\prismcafo{} achieves +15\% over the best baseline, while ViT-B/16+\prismcafo{} lifts Dairy by +31\% and Beef by +42\%. RemoteCLIP again shows consistent but smaller gains. 

In summary, \prismcafo{} consistently improves performance, with the largest relative boosts in Dairy and Beef. CNNs see the largest relative improvements, while transformers reach the highest absolute scores. Overall, the best model (Swin-B+\prismcafo{}) outperforms the best baseline (DINOv2 ViT-B/16) by +4\% under random splits and +15\% under spatial splits.

\begin{table}[h]
\centering
\setlength{\tabcolsep}{6pt}
\begin{tabular}{cccc|c}
\toprule
MGSA & MAP & SIM & PFV & macro-F1 \\
\midrule
\xmark &  \xmark &  \xmark & \xmark &   0.718 \\
\xmark &  \cmark  & \cmark & \cmark &   0.792\\
\xmark &  \cmark  & \cmark & \xmark  &  0.793\\
\cmark &  \xmark  & \xmark & \xmark  &  0.803\\
\cmark &  \xmark  & \cmark & \cmark  &  0.830\\
\cmark &  \cmark  & \cmark & \cmark  &  \textbf{0.841}\\
\bottomrule
\end{tabular}
\caption{Ablation study of core modules. MGSA: Mask-Guided Spatial Attention, MAP: Masked Attention Pooling, SIM: Synthetic Infrastructure Masks, PFV: Prior Feature Vector. (\cmark\ ) indicates inclusion and (\xmark\ ) exclusion of a component. Note that to conduct this, we have used EfficientNet-B0.}
\label{tab:ablation}
\end{table}


\subsection{Evaluating Significance of Components}

All our studies in this section are based on the  EfficientNet-B0 backbone.

\noindent\textbf{Ablation Study.}
To assess the contribution of each component in our framework, we
perform an ablation study. Table~\ref{tab:ablation} reports macro-F1 scores with different module combinations.  The baseline without any component achieves 0.718 macro-F1. Adding MGSA alone boosts performance by +12\%. Incorporating SIM and MAP together gives +10\%, while further adding PFV yields stable gains. The best performance (0.841) arises when all four modules are combined, improving macro-F1 by +17\% over the baseline. This demonstrates the importance of each component.
\\
\noindent\textbf{Importance of extracted features and domain priors.} 
We use gradient-activation~\cite{ancona2017towards,wang2021feature} analysis for this purpose. Let $\mathbf{x}_{\text{in}}\!\in\!\mathbb{R}^{D}$ be the vector entering the final linear classifier. For the predicted class $c^\star\in {\text{\{beef, poultry, swine, dairy\}}}$, we compute
\[
\mathbf{g} = \frac{\partial y_{c^\star}}{\partial \mathbf{x}_{\text{in}}}, 
\quad
\mathbf{s} = |\mathbf{g}| \circ |\mathbf{x}_{in}|,
\]
where $\circ$ denotes element-wise (Hadamard~\cite{horn1990hadamard}) multiplication. 
We use absolute values to remove sign ambiguity, as the gradient can be positive or negative. This score reflects how much each feature contributes to the prediction, combining its activation strength with the model’s sensitivity.

\begin{figure}
    \centering
    \includegraphics[width=\linewidth]{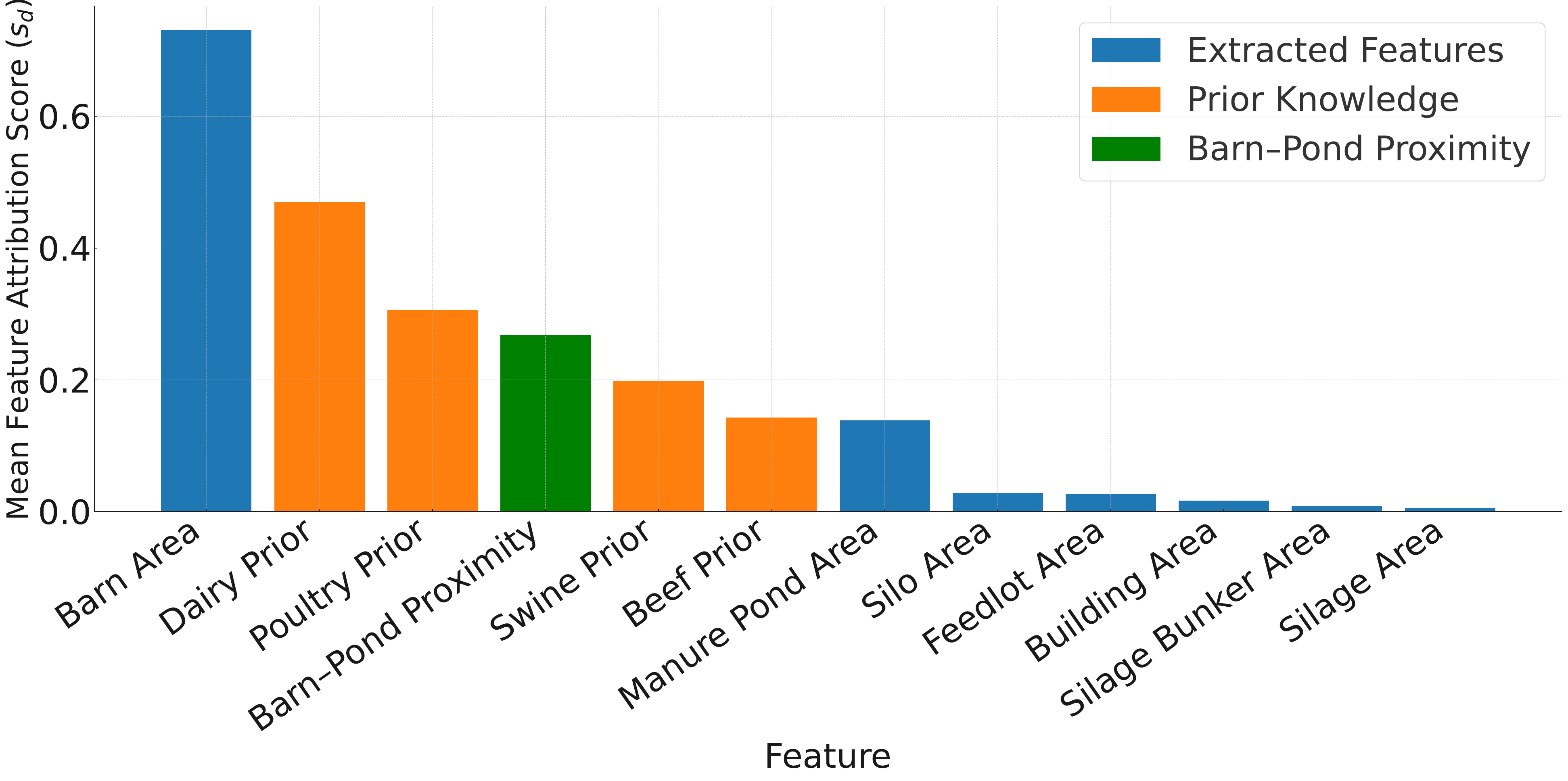}
    \caption{Feature importance scores using gradient-activation analysis ($|\nabla \cdot \text{act}|$) for livestock classification. Barn area dominates (0.74), followed by domain priors (0.15--0.47) and barn-pond proximity (0.27). Blue: extracted features, orange: prior knowledge, green: proximity measures.}
    \label{fig:attribute-score}
\end{figure}

\begin{figure}
    \centering
    \includegraphics[width=\linewidth]{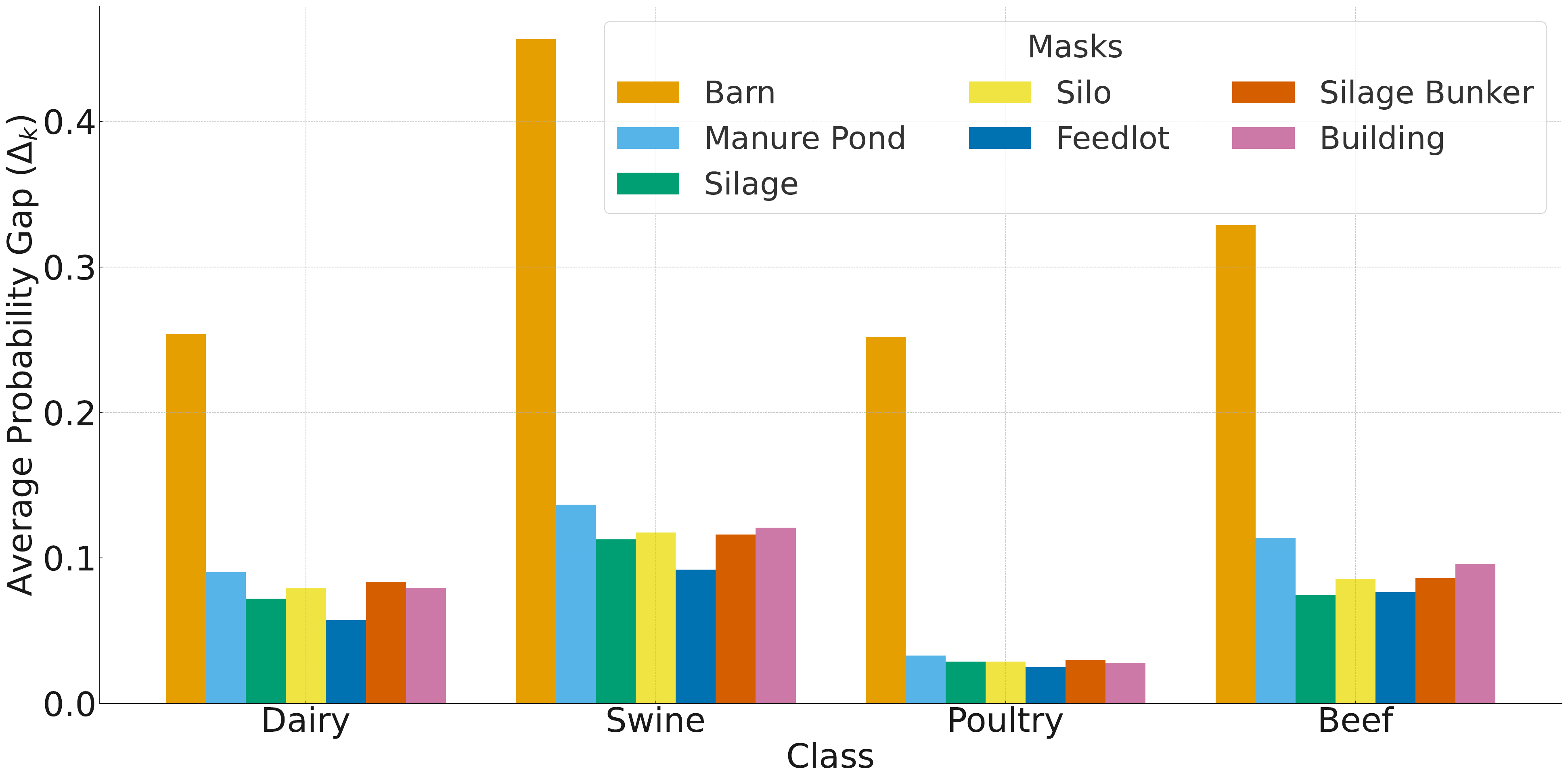}
    \caption{Probability-drop (\(\Delta_k\)) analysis of infrastructure masks. Barns dominate across all classes (0.45 swine, 0.33 beef, and 0.25 dairy and poultry), with manure ponds adding signals for swine and beef. Other components remain comparatively small.}
    \label{fig:probability-gap}
\end{figure}

Based on the gradient-activation analysis, barn area exhibits the highest importance score (0.74), outweighing all other features (Fig.~\ref{fig:attribute-score}). This dominance is because barn size directly determines livestock capacity, serving as the primary discriminator between production types. Domain knowledge features show importance scores of 0.47 (dairy), 0.31 (poultry), 0.20 (swine), and 0.15 (beef) which provides regional agricultural context. Barn-pond proximity contributes meaningfully (0.27), reflecting species-specific water access requirements—dairy operations often need nearby water sources, while poultry benefits from greater separation to reduce contamination risk. Other features like silo area are discriminative for dairy operations. This analysis confirms that physical infrastructure characteristics, particularly barn size, constitute the most reliable classification signals, while domain knowledge and spatial relationships provide essential contextual information for accurate agricultural land use classification.
\\
\noindent\textbf{Importance of Infrastructure Masks.} 
In the supplement, statistics of the candidate 
bounding boxes are plotted for the various infrastructure types.
While some structures such as feedlot~(beef) and silo~(swine) seem to
be discriminative objects~(potentially assisting in classification),
there is a  significant intersection in infrastructure sets across
livestock types making it non-trivial to analyze the importance of a
infrastructure type from statistics alone.
In order to systematically analyze any such significance, we attribute importance to each component channel $M_k \in [0,1]^{H\times W}$, corresponding to barns, ponds, silos, and other structures using probability drop~($\Delta$)~\cite{fong2017interpretable,ramaswamy2020ablation}. For predicted class $c^\star$, we measure the change in probability when channel $k$ is masked out:
\[
\begin{aligned}
\Delta_{-k}^{+} &= \big[p_{c^\star}(x,M)-p_{c^\star}(x,M_{-k})\big]_+, && M_{-k}:\; M_k \equiv 0,\\
\Delta_{+k}^{+} &= \big[p_{c^\star}(x,M)-p_{c^\star}(x,M_{+k})\big]_+, && M_{+k}:\; M_k \equiv 1,\\
\Delta_k &= \tfrac{1}{2}\big(\Delta_{-k}^{+}+\Delta_{+k}^{+}\big), && [a]_+=\max(a,0).
\end{aligned}
\]
The larger the $\Delta_k$, the higher the significance; removing the~$k$th channel significantly reduces confidence.
The probability-drop analysis reveals that all infrastructure masks contribute meaningfully to livestock facility classification~(Fig.~\ref{fig:probability-gap}). Barns emerge as the type of highest
importance across all facility types which validates their role as primary housing structures for livestock operation and detection.
Manure ponds provide substantial discriminative power for swine and beef operations, confirming these facilities' reliance on lagoon-based waste management systems. Dairy CAFOs benefit from diverse infrastructure, including silos, silage bunkers, and feedlots. This reflects complex feed storage requirements, while swine operations utilize multiple components with consistent importance. 
Poultry farms' heavy dependence on barns is consistent with their characteristic footprints (i.e., long and parallel structures dominating commercial operations). 

\section{Conclusion}
We propose a detector-anchored pipeline that bootstraps YOLO with SAM2 masks and geometric filters to generate reliable infrastructure instances, enabling a mask-guided classifier that matches vision-only baselines while offering greater interpretability through object and relation-level reasoning.

\noindent\paragraph{Generalization and Scalability.} Although our experiments focus on CAFO identification, PRISM-CAFO pipeline can be adapted to a wide range of land features such as energy facilities, warehouses, and other industrial sites. Adapting the method requires only a modest bootstrapping phase in which a small set of domain-specific components are annotated and the same containment, shape, and proximity filters are retuned for the new object vocabulary. Our structured descriptors—area ratios, pairwise spatial relationships, and region-level priors—are likewise transferable and can be redefined over any set of infrastructure elements. Because the mask-guided attention and mask-aware pooling operate purely on predicted instance masks, the approach naturally accommodates new satellite or aerial imagery sources and evolving infrastructure layouts. In this way, the methodology scales beyond livestock facilities and provides a general template for infrastructure-conditioned segmentation and classification in remotely sensed imagery.

\noindent\paragraph{Limitations.} While our results demonstrate strong gains, limitations remain. Transformer-based vision models underperform in this setting, highlighting the need for specialized designs that better exploit structured infrastructure cues. Moreover, our spatial split evaluation suggests that models require at least a few positive examples from new regions to generalize effectively, pointing to semi-supervised or active learning as promising directions. Beyond CAFOs, our infrastructure-first methodology could be extended to other domains such as energy facilities, warehouses, or data centers, where fine-grained infrastructure composition is equally critical for environmental monitoring and policy assessment. The generated infrastructure candidates can also strengthen open-vocabulary methods by grounding prompts with precise object instances, guide multimodal LLMs with structured descriptors, and support environmental monitoring. For example, barns indicate livestock density and ammonia emissions, lagoons signal nutrient runoff risk, and feedlots reveal land use intensity and methane generation—linking infrastructure-level vision to measurable environmental impacts.

\noindent\paragraph{Future Directions.} Future work includes unifying the detection, segmentation, and classification stages into an end-to-end differentiable pipeline; augmenting our method with multi-temporal observations to capture operational dynamics; and integrating outputs with downstream regulatory, epidemiological, and environmental models. Another direction is incorporating vision–language foundation models to enable textual explanations or open-vocabulary infrastructure tags. Finally, extending our approach to global imagery and exploring semi-supervised and active learning frameworks will further enhance scalability and reduce annotation burden.

\section*{Acknowledgments} This work was partially supported by NSF and USDA-NIFA under the AI Institute: Agricultural AI for Transforming Workforce and Decision Support (AgAID) award No. 2021-67021-35344, University of Virginia Strategic Investment Fund award number SIF160 and PGCoE CDC-RFA-CK22-2204.
{\small
\bibliographystyle{ieee_fullname}
\bibliography{reference}

@inproceedings{ramaswamy2020ablation,
  title={Ablation-cam: Visual explanations for deep convolutional network via gradient-free localization},
  author={Ramaswamy, Harish Guruprasad and others},
  booktitle={proceedings of the IEEE/CVF winter conference on applications of computer vision},
  pages={983--991},
  year={2020}
}

@article{bakshi2023near,
  title={Near-linear time algorithm for the chamfer distance},
  author={Bakshi, Ainesh and Indyk, Piotr and Jayaram, Rajesh and Silwal, Sandeep and Waingarten, Erik},
  journal={Advances in Neural Information Processing Systems},
  volume={36},
  pages={66833--66844},
  year={2023}
}

@inproceedings{horn1990hadamard,
  title={The {H}adamard product},
  author={Horn, Roger A},
  booktitle={Proc. symp. appl. math},
  volume={40},
  pages={87--169},
  year={1990}
}

@inproceedings{fong2017interpretable,
  title={Interpretable explanations of black boxes by meaningful perturbation},
  author={Fong, Ruth C and Vedaldi, Andrea},
  booktitle={Proceedings of the IEEE international conference on computer vision},
  pages={3429--3437},
  year={2017}
}

@inproceedings{wang2021feature,
  title={Feature importance-aware transferable adversarial attacks},
  author={Wang, Zhibo and Guo, Hengchang and Zhang, Zhifei and Liu, Wenxin and Qin, Zhan and Ren, Kui},
  booktitle={Proceedings of the IEEE/CVF international conference on computer vision},
  pages={7639--7648},
  year={2021}
}

@article{ancona2017towards,
  title={Towards better understanding of gradient-based attribution methods for deep neural networks},
  author={Ancona, Marco and Ceolini, Enea and {\"O}ztireli, Cengiz and Gross, Markus},
  journal={arXiv preprint arXiv:1711.06104},
  year={2017}
}

@misc{nlcd,
  author = {Multi-Resolution Land Characteristics Consortium (MRLC)},
  title = {{National Land Cover Database (NLCD)}},
  year = {2021},
  url = {https://www.mrlc.gov/},
  note = {Accessed: 2024-03-01}
}

@misc{cafomaps,
  author       = {{Department of Geographical and Sustainability Sciences, University of Iowa}},
  title        = {CAFOMaps: Concentrated Animal Feeding Operations in the {U}nited {S}tates},
  year         = {2025},
  howpublished = {\url{https://www.cafomaps.org/}},
  note         = {Accessed: 2025-05-16}
}

@article{oquab2023dinov2,
  title={{DINOv2}: Learning Robust Visual Features without Supervision},
  author={Oquab, Maxime and Darcet, Thomas and Moutakanni, Theo and Ram{\'e}, Alexandre and Haziza, Daniel and Su{\'a}rez, Jorge and Szafraniec, Marc and Kalantidis, Yannis and Elkabetz, Yair and Cord, Matthieu and others},
  journal={arXiv:2304.07193},
  year={2023}
}

@article{heikel2022tfidf,
  author    = {Edvard Heikel and Leonardo Espinosa{-}Leal},
  title     = {Indoor Scene Recognition via Object Detection and {TF-IDF}},
  journal   = {J. Imaging},
  volume    = {8},
  number    = {8},
  pages     = {209},
  year      = {2022},
  doi       = {10.3390/jimaging8080209},
  url       = {https://www.mdpi.com/2313-433X/8/8/209}
}

@article{song2024interobject,
  author    = {Chenyu Song and Yuzhe Yang and others},
  title     = {Inter-object Discriminative Graph Modeling for Indoor Scene Recognition},
  journal   = {Knowledge-Based Systems},
  volume    = {298},
  pages     = {112371},
  year      = {2024},
  doi       = {10.1016/j.knosys.2024.112371},
  url       = {https://www.sciencedirect.com/science/article/pii/S0950705124010050}
}

@inproceedings{li2024ovsgg,
  author    = {Rui Li and others},
  title     = {From Pixels to Graphs: Open-Vocabulary Scene Graph Generation with Vision-Language Models},
  booktitle = {CVPR},
  year      = {2024},
  url       = {https://openaccess.thecvf.com/content/CVPR2024/html/Li_From_Pixels_to_Graphs_Open-Vocabulary_Scene_Graph_Generation_with_Vision-Language_CVPR_2024_paper.html}
}

@inproceedings{li2024predicate,
  author    = {Jiaxin Li and others},
  title     = {Leveraging Predicate and Triplet Learning for Scene Graph Generation},
  booktitle = {CVPR},
  year      = {2024},
  url       = {https://openaccess.thecvf.com/content/CVPR2024/html/Li_Leveraging_Predicate_and_Triplet_Learning_for_Scene_Graph_Generation_CVPR_2024_paper.html}
}

@article{cheng2023sammed,
  author    = {Cheng, K. and others},
  title     = {{SAM-Med}: Adapting Segment Anything Model for Medical Image Segmentation},
  journal   = {arXiv:2304.12620},
  year      = {2023}
}

@article{ke2024universal,
  author    = {Ke, Lei and others},
  title     = {Universal Segmentation at Any Granularity with Mask Prompting},
  journal   = {CVPR},
  year      = {2024}
}

@inproceedings{radford2021clip,
  author    = {Radford, Alec and others},
  title     = {Learning Transferable Visual Models From Natural Language Supervision},
  booktitle = {ICML},
  year      = {2021}
}

@inproceedings{yuan2021florence,
  author    = {Yuan, Li and others},
  title     = {Florence: A New Foundation Model for Computer Vision},
  booktitle = {CVPR},
  year      = {2021}
}

@article{jocher2023yolov8,
  author    = {Glenn Jocher and others},
  title     = {{YOLOv8}: Cutting-edge Object Detection},
  journal   = {ultralytics.com},
  year      = {2023}
}

@article{ding2024specializeddet,
  author    = {Ding, Y. and others},
  title     = {Specializing Generic Object Detectors to Domain-specific Categories via Weak Supervision},
  journal   = {Pattern Recognition},
  year      = {2024}
}

@inproceedings{kirillov2023sam,
  title        = {Segment Anything},
  author       = {Kirillov, Alexander and Mintun, Eric and Ravi, Nikhila and Mao, Hanzi and Rolland, Chloe and Gustafson, Laura and Xiao, Tete and Whitehead, Spencer and Berg, Alexander C. and Lo, Wan-Yen and Doll{\'a}r, Piotr and Girshick, Ross},
  booktitle    = {ICCV},
  year         = {2023},
  url          = {https://arxiv.org/abs/2304.02643}
}

@inproceedings{minderer2023owlv2,
  title        = {Scaling Open-Vocabulary Object Detection},
  author       = {Minderer, Matthias and Gritsenko, Alexey and Houlsby, Neil and others},
  booktitle    = {NeurIPS},
  year         = {2023},
  url          = {https://arxiv.org/abs/2306.09683}
}

@inproceedings{liang2023ovseg,
  title        = {Open-Vocabulary Semantic Segmentation with {M}ask-adapted {CLIP}},
  author       = {Liang, Feng and Wu, Bichen and Dai, Xiaoliang and Li, Kunpeng and Zhao, Yinan and Zhang, Hang and Zhang, Peizhao and Vajda, Peter and Marculescu, Diana},
  booktitle    = {CVPR},
  year         = {2023},
  url          = {https://arxiv.org/abs/2210.04150}
}

@inproceedings{xiao2024florence2,
  title        = {Florence-2: Advancing a Unified Representation for a Variety of Vision Tasks},
  author       = {Xiao, Bin and Wu, Haiping and Xu, Weijian and Dai, Xiyang and Hu, Houdong and Lu, Yumao and Zeng, Michael and Liu, Ce and Yuan, Lu},
  booktitle    = {CVPR},
  year         = {2024},
  url          = {https://openaccess.thecvf.com/content/CVPR2024/papers/Xiao_Florence-2_Advancing_a_Unified_Representation_for_a_Variety_of_Vision_CVPR_2024_paper.pdf}
}

@inproceedings{lee2013pseudolabel,
  title        = {Pseudo-Label: The Simple and Efficient Semi-Supervised Learning Method for Deep Neural Networks},
  author       = {Lee, Dong-Hyun},
  booktitle    = {ICML Workshop on Challenges in Representation Learning (WREPL)},
  year         = {2013},
  url          = {https://storage.googleapis.com/kaggle-forum-message-attachments/7371/pseudo_label_draft.pdf}
}

@misc{agcensus2022,
	author = {{U}{S}{D}{A} - {N}ational {A}gricultural {S}tatistics {S}ervice},
	title = {{C}ensus of {A}griculture},
	howpublished = {\url{https://www.nass.usda.gov/AgCensus/}},
	year = {2022},
	note = {[Accessed 03-Jan-2023]},
}

@inproceedings{xie2020noisystudent,
  title        = {Self-Training with Noisy Student Improves {I}mage{N}et Classification},
  author       = {Xie, Qizhe and Luong, Minh-Thang and Hovy, Eduard and Le, Quoc V.},
  booktitle    = {CVPR},
  year         = {2020},
  url          = {https://openaccess.thecvf.com/content_CVPR_2020/papers/Xie_Self-Training_With_Noisy_Student_Improves_ImageNet_Classification_CVPR_2020_paper.pdf}
}

@article{exploiting2024objseg,
  author    = {Ricardo Pereira and Luís Garrote and Tiago Barros and Ana Lopes and Urbano J. Nunes},
  title     = {Exploiting Object-based and Segmentation-informed Representations for Indoor Scene Classification},
  journal   = {arXiv:2404.07739},
  year      = {2024},
  url       = {https://arxiv.org/html/2404.07739v1}
}

@article{zhou2025hierarchical,
  author    = {J. Zhou and others},
  title     = {Scene Categorization by {H}essian-regularized Active Multi-channel Perceptual Features},
  journal   = {Scientific Reports},
  volume    = {15},
  number    = {1},
  year      = {2025},
  doi       = {10.1038/s41598-024-84181-x},
  url       = {https://www.nature.com/articles/s41598-024-84181-x}
}

@article{ravi2024sam2,
  title={{SAM} 2: Segment Anything in Images and Videos},
  author={Ravi, Nikhila and Gabeur, Valentin and Hu, Yuan-Ting and Hu, Ronghang and Ryali, Chaitanya and Ma, Tengyu and Khedr, Haitham and R{\"a}dle, Roman and Rolland, Chloe and Gustafson, Laura and Mintun, Eric and Pan, Junting and Alwala, Kalyan Vasudev and Carion, Nicolas and Wu, Chao-Yuan and Girshick, Ross and Doll{\'a}r, Piotr and Feichtenhofer, Christoph},
  journal={arXiv:2408.00714},
  url={https://arxiv.org/abs/2408.00714},
  year={2024}
}

@misc{USGS_NAIP,
  author = {{U.S. Department of Agriculture}},
  title = {{National Agriculture Imagery Program (NAIP)}},
  howpublished = {{USGS Science Data Catalog}},
  year = {2023},
  url = {{https://naip-usdaonline.hub.arcgis.com/}},
  note = {{Dataset acquired by the U.S. Department of Agriculture (USDA)}}
}

@article{moses2017industrial,
  title={Industrial animal agriculture in the {U}nited {S}tates: Concentrated animal feeding operations (CAFOs)},
  author={Moses, Aurora and Tomaselli, Paige},
  journal={International farm animal, wildlife and food safety law},
  pages={185--214},
  year={2017},
  publisher={Springer}
}

@techreport{ehrenpreis2021nmcafo,
  author      = {Vanessa Ehrenpreis and Marshall Worsham and Nick Clarke and Adam Buchholz},
  title       = {Using Machine Learning to Map Concentrated Animal Feeding Operations in {N}ew {M}exico},
  institution = {Mapping for Environmental Justice},
  year        = {2021},
  month       = {January},
  note        = {Report prepared for the McGovern Foundation},
  url         = {https://mappingforej.studentorg.berkeley.edu/wp-content/uploads/2022/03/NM-CAFO-Report.pdf}
}

@article{handan2019deep,
  title={Deep learning to map concentrated animal feeding operations},
  author={Handan-Nader, Cassandra and Ho, Daniel E},
  journal={Nature Sustainability},
  volume={2},
  number={4},
  pages={298--306},
  year={2019},
  publisher={Nature Publishing Group UK London}
}

@article{zhu2022meter,
  title={{METER-ML}: a multi-sensor earth observation benchmark for automated methane source mapping},
  author={Zhu, Bryan and Lui, Nicholas and Irvin, Jeremy and Le, Jimmy and Tadwalkar, Sahil and Wang, Chenghao and Ouyang, Zutao and Liu, Frankie Y and Ng, Andrew Y and Jackson, Robert B},
  journal={arXiv preprint arXiv:2207.11166},
  year={2022}
}

@article{nguyen2025emergence,
  title={Emergence and interstate spread of highly pathogenic avian influenza A (H5N1) in dairy cattle in the {U}nited {S}tates},
  author={Nguyen, Thao-Quyen and Hutter, Carl R and Markin, Alexey and Thomas, Megan and Lantz, Kristina and Killian, Mary Lea and Janzen, Garrett M and Vijendran, Sriram and Wagle, Sanket and Inderski, Blake and others},
  journal={Science},
  volume={388},
  number={6745},
  pages={eadq0900},
  year={2025},
  publisher={American Association for the Advancement of Science}
}

@article{saha2025machine,
  title={Machine learning-based identification of animal feeding operations in the {U}nited {S}tates on a parcel-scale},
  author={Saha, Arghajeet and Rashid, Barira and Liu, Ting and Miralha, Lorrayne and Muenich, Rebecca L},
  journal={Science of The Total Environment},
  volume={960},
  pages={178312},
  year={2025},
  publisher={Elsevier}
}

@article{adiga2024high,
  title={A High-Resolution, US-scale Digital Similar of Interacting Livestock, Wild Birds, and Human Ecosystems with Applications to Multi-host Epidemic Spread},
  author={Adiga, Abhijin and Chopra, Ayush and Wilson, Mandy L and Ravi, SS and Xie, Dawen and Swarup, Samarth and Lewis, Bryan and Raskar, Ramesh and Marathe, Madhav V},
  journal={arXiv preprint arXiv:2411.01386},
  year={2024}
}

@article{humphreys2020waterfowl,
    author = "Humphreys, John M and Ramey, Andrew M and Douglas, David C and Mullinax, Jennifer M and Soos, Catherine and Link, Paul and Walther, Patrick and Prosser, Diann J",
    title = "Waterfowl occurrence and residence time as indicators of H5 and H7 avian influenza in {N}orth {A}merican Poultry",
    journal = "Scientific Reports",
    volume = "10",
    number = "1",
    pages = "2592",
    year = "2020",
    publisher = "Nature Publishing Group UK London",
    original_key = "humphreys2020waterfowl"
}

@article{prosser2024using,
    author = "Prosser, Diann J and Kent, Cody M and Sullivan, Jeffery D and Patyk, Kelly A and McCool, Mary-Jane and Torchetti, Mia Kim and Lantz, Kristina and Mullinax, Jennifer M",
    title = "Using an adaptive modeling framework to identify avian influenza spillover risk at the wild-domestic interface",
    journal = "Scientific Reports",
    volume = "14",
    number = "1",
    pages = "14199",
    year = "2024",
    publisher = "Nature Publishing Group UK London",
    original_key = "prosser2024using"
}

@ARTICLE{9832662,
  author={Robinson, Caleb and Chugg, Ben and Anderson, Brandon and Ferres, Juan M. Lavista and Ho, Daniel E.},
  journal={IEEE Journal of Selected Topics in Applied Earth Observations and Remote Sensing}, 
  title={Mapping Industrial Poultry Operations at Scale With Deep Learning and Aerial Imagery}, 
  year={2022},
  volume={15},
  number={},
  pages={7458-7471},
  doi={10.1109/JSTARS.2022.3191544}}

@article{CHUGG2021102463,
title = {Enhancing environmental enforcement with near real-time monitoring: Likelihood-based detection of structural expansion of intensive livestock farms},
journal = {International Journal of Applied Earth Observation and Geoinformation},
volume = {103},
pages = {102463},
year = {2021},
issn = {1569-8432},
doi = {https://doi.org/10.1016/j.jag.2021.102463},
url = {https://www.sciencedirect.com/science/article/pii/S0303243421001707},
author = {Ben Chugg and Brandon Anderson and Seiji Eicher and Sandy Lee and Daniel E. Ho}
}

@book{GurianSherman2008,
  author    = {D. Gurian-Sherman},
  title     = {{CAFOs Uncovered: The Untold Costs of Confined Animal Feeding Operations}},
  year      = {2008},
  publisher = {Union of Concerned Scientists}
}

@techreport{Hribar2010,
  author      = {C. Hribar},
  title       = {{Understanding Concentrated Animal Feeding Operations and Their Impact on Communities}},
  institution = {National Association of Local Boards of Health},
  year        = {2010}
}

@incollection{HandanNader2021,
  author    = {C. Handan-Nader and D. E. Ho and L. Y. Liu},
  title     = {{Deep Learning with Satellite Imagery to Enhance Environmental Enforcement}},
  booktitle = {{Data Science Applied to Sustainability Analysis}},
  pages     = {205--228},
  publisher = {Elsevier},
  year      = {2021}
}

@inproceedings{selvaraju2017grad,
  title={Grad-CAM: Visual explanations from deep networks via gradient-based localization},
  author={Selvaraju, Ramprasaath R. and Cogswell, Michael and Das, Abhishek and Vedantam, Ramakrishna and Parikh, Devi and Batra, Dhruv},
  booktitle={Proceedings of the IEEE International Conference on Computer Vision (ICCV)},
  pages={618--626},
  year={2017}
}

@article{zou2023segment,
  title={Segment and Track Anything},
  author={Zou, Yu and Yang, Linjie and Wang, Zhaoyang and Huang, Jia-Bin},
  journal={arXiv preprint arXiv:2306.00989},
  year={2023}
}

@misc{usda_naip,
  author       = {{U.S. Department of Agriculture}},
  title        = {{National Agriculture Imagery Program (NAIP)}},
  year         = {2023},
  note         = {\url{https://www.fsa.usda.gov/programs-and-services/aerial-photography/imagery-programs/naip-imagery/}},
  howpublished = {\url{https://www.fsa.usda.gov/programs-and-services/aerial-photography/imagery-programs/naip-imagery/}},
  institution  = {USDA Farm Service Agency},
}

@misc{NYcafo,
  author       = {{New York Department of State}},
  title        = {{New York CAFO Dataset}},
  year         = {2024},
  howpublished = {\url{https://opdgig.dos.ny.gov/datasets/a9a8eaed80864ab98680899ecdbc1c50/explore?location=42.664852\%2C-76.586900\%2C7.22}},
  note         = {Accessed: 2024-05-15}
}

@misc{MNcafo,
  author       = {Minnesota Geospatial Commons},
  title        = {Minnesota CAFO Locations (Layer 0)},
  year         = {2024},
  howpublished = {\url{https://www.arcgis.com/home/item.html?id=6d119156229d4e908e22f027bdaee6be\&sublayer=0}},
  note         = {Accessed: 2024-05-15}
}

@misc{MIcafo,
  author       = {{Michigan EGLE GIS Hub}},
  title        = {{CAFO Locations -- Michigan Department of Environment, Great Lakes, and Energy}},
  year         = {2024},
  howpublished = {\url{https://gis-egle.hub.arcgis.com/datasets/f0843875e5874d04b06396de8200cf75/explore?location=43.005451\%2C-83.963570\%2C6.10}},
  note         = {Accessed: 2024-05-15}
}

@misc{MDcafo,
  author       = {Maryland Department of the Environment},
  title        = {Animal Feeding Operations Map},
  year         = {2024},
  howpublished = {\url{https://catalog.data.gov/dataset/maryland-department-of-the-environment-lma-resource-management-program-animal-feeding-oper-5fb42}},
  note         = {Accessed: 2024-05-15}
}

@misc{IAcafo,
  author       = {Iowa Geodata Portal},
  title        = {Iowa Animal Feeding Operations GIS Data},
  year         = {2024},
  howpublished = {\url{https://geodata.iowa.gov/documents/abfbd972640d4e87b6c48dc669775767/about}},
  note         = {Accessed: 2024-05-15}
}

@misc{INcafo,
  author       = {IndianaMap},
  title        = {Confined Feeding Operations},
  year         = {2024},
  howpublished = {\url{https://www.indianamap.org/datasets/INMap::confined-feeding-operations/about}},
  note         = {Accessed: 2024-05-15}
}

@misc{DEcafo,
  author       = {Delaware Department of Natural Resources},
  title        = {Delaware CAFO Map Viewer},
  year         = {2024},
  howpublished = {\url{https://experience.arcgis.com/experience/c6749f8b31d143cbb38a26fc1b89a2be/page/Delaware}},
  note         = {Accessed: 2024-05-15}
}

@inproceedings{he2016resnet,
  title     = {Deep Residual Learning for Image Recognition},
  author    = {He, Kaiming and Zhang, Xiangyu and Ren, Shaoqing and Sun, Jian},
  booktitle = {CVPR},
  year      = {2016},
  pages     = {770--778},
  doi       = {10.1109/CVPR.2016.90},
  url       = {https://openaccess.thecvf.com/content_cvpr_2016/html/He_Deep_Residual_Learning_CVPR_2016_paper.html}
}

@inproceedings{tan2019efficientnet,
  title     = {EfficientNet: Rethinking Model Scaling for Convolutional Neural Networks},
  author    = {Tan, Mingxing and Le, Quoc},
  booktitle = {ICML},
  year      = {2019},
  pages     = {6105--6114},
  publisher = {PMLR},
  url       = {https://proceedings.mlr.press/v97/tan19a.html}
}

@inproceedings{dosovitskiy2021vit,
  title     = {An Image is Worth 16x16 Words: Transformers for Image Recognition at Scale},
  author    = {Dosovitskiy, Alexey and Beyer, Lucas and Kolesnikov, Alexander and Weissenborn, Dirk and Zhai, Xiaohua and Unterthiner, Thomas and Dehghani, Mostafa and Minderer, Matthias and Heigold, Georg and Gelly, Sylvain and Uszkoreit, Jakob and Houlsby, Neil},
  booktitle = {ICLR},
  year      = {2021},
  url       = {https://openreview.net/forum?id=YicbFdNTTy}
}

@inproceedings{liu2021swin,
  title     = {Swin Transformer: Hierarchical Vision Transformer using Shifted Windows},
  author    = {Liu, Ze and Lin, Yutong and Cao, Yue and Hu, Han and Wei, Yixuan and Zhang, Zheng and Lin, Stephen and Guo, Baining},
  booktitle = {ICCV},
  year      = {2021},
  pages     = {10012--10022},
  url       = {https://openaccess.thecvf.com/content/ICCV2021/html/Liu_Swin_Transformer_Hierarchical_Vision_Transformer_Using_Shifted_Windows_ICCV_2021_paper.html}
}

@article{liu2023remoteclip,
  title   = {RemoteCLIP: A Vision-Language Foundation Model for Remote Sensing},
  author  = {Liu, Fan and Chen, Delong and Zhang, Zhitong and Jiao, Licheng and et al.},
  journal = {arXiv preprint arXiv:2306.11029},
  year    = {2023},
  url     = {https://arxiv.org/abs/2306.11029}
}
}

\clearpage
\appendix
\onecolumn

\section{Data Sources}
\label{sec:data_sources}
\noindent Here, we provide a summary of data sources and acquisition. Details are in the supplement. 
Table~\ref{tab:cafo-sources} lists all data sources.
\\
\noindent \textbf{Satellite Imagery.}
We leverage aerial imagery from the National Agriculture Imagery Program (NAIP), accessed via the 
Microsoft  Planetary Computer. For each geolocated CAFO point, the nearest cloud-free NAIP image 
is queried and downloaded using the STAC API. 
For this experimentation, we collected most recent~(2023) data for the studied states.
\\
\noindent \textbf{CAFO datasets.}
The references to all data sources are in Table~\ref{tab:cafo-sources}. We obtained data from
multiple sources.
Department of 
Geographical and Sustainability Sciences of IOWA (denoted as IOWA-CAFO Inventory) and animal 
feeding operation reports from states where available (denoted as State-CAFO Inventory). 
IOWA-CAFO inventory aggregates CAFO facility data from state environmental agencies across nine 
southeastern US states. Data is collected from permit databases, nutrient management plans, and 
agency inspections. Each record includes geolocation, animal type (poultry, swine, beef, dairy), 
and manure management details.
Several US states publish CAFO reports curated by official state agencies using permit records, 
inspection data, and self-reported nutrient management plans.  See Table~\ref{tab:cafo-sources} for
the states and the sources~(row 3).
\\
\noindent \textbf{Land Use Masks.}
We utilize national-scale raster products to identify and contextualize  agricultural areas. The MRLC National Land Cover 
Database~(NLCD) offers 30m-resolution land cover 
classifications across 16 categories, including cultivated cropland, grassland, barren land, and pasture~(see Table~\ref{tab:cafo-sources}). This dataset is further used to generate stratified negative samples based on land cover types and spatial extents.

\section{Additional Results}
\begin{figure*}[htbp]
    \centering
    \includegraphics[width=0.48\linewidth]{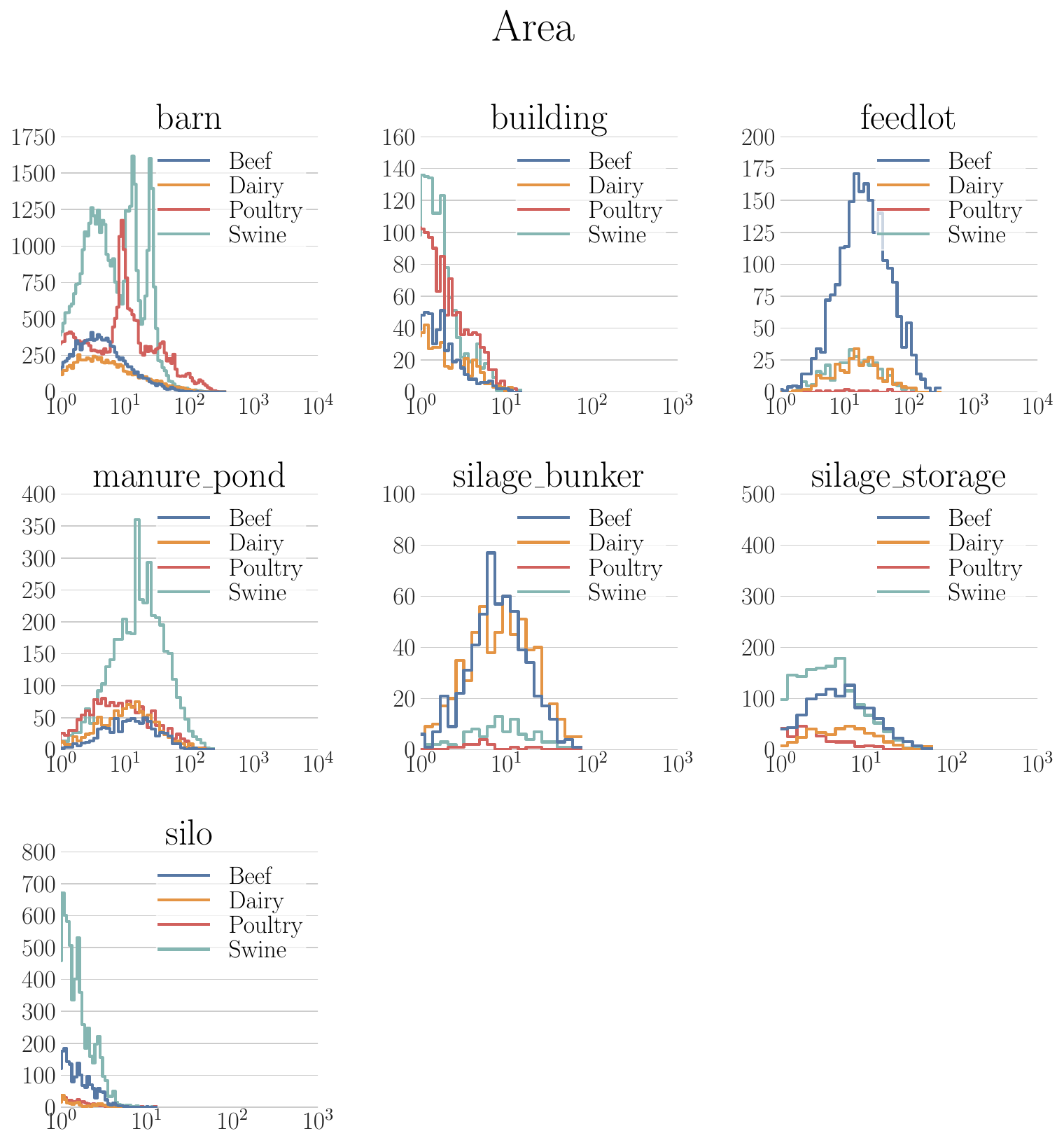}
    \includegraphics[width=0.48\linewidth]{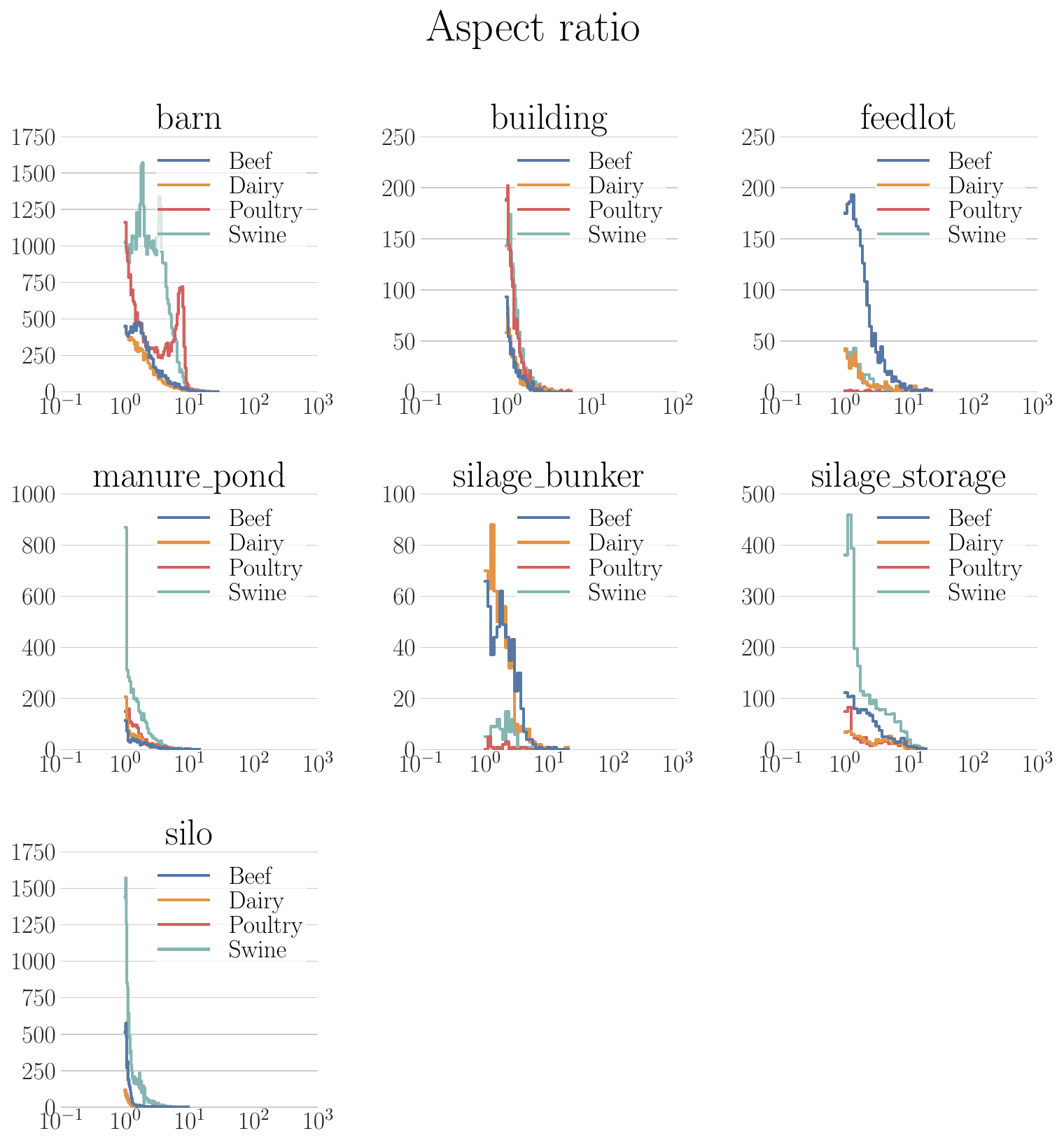}
    \includegraphics[width=0.48\linewidth]{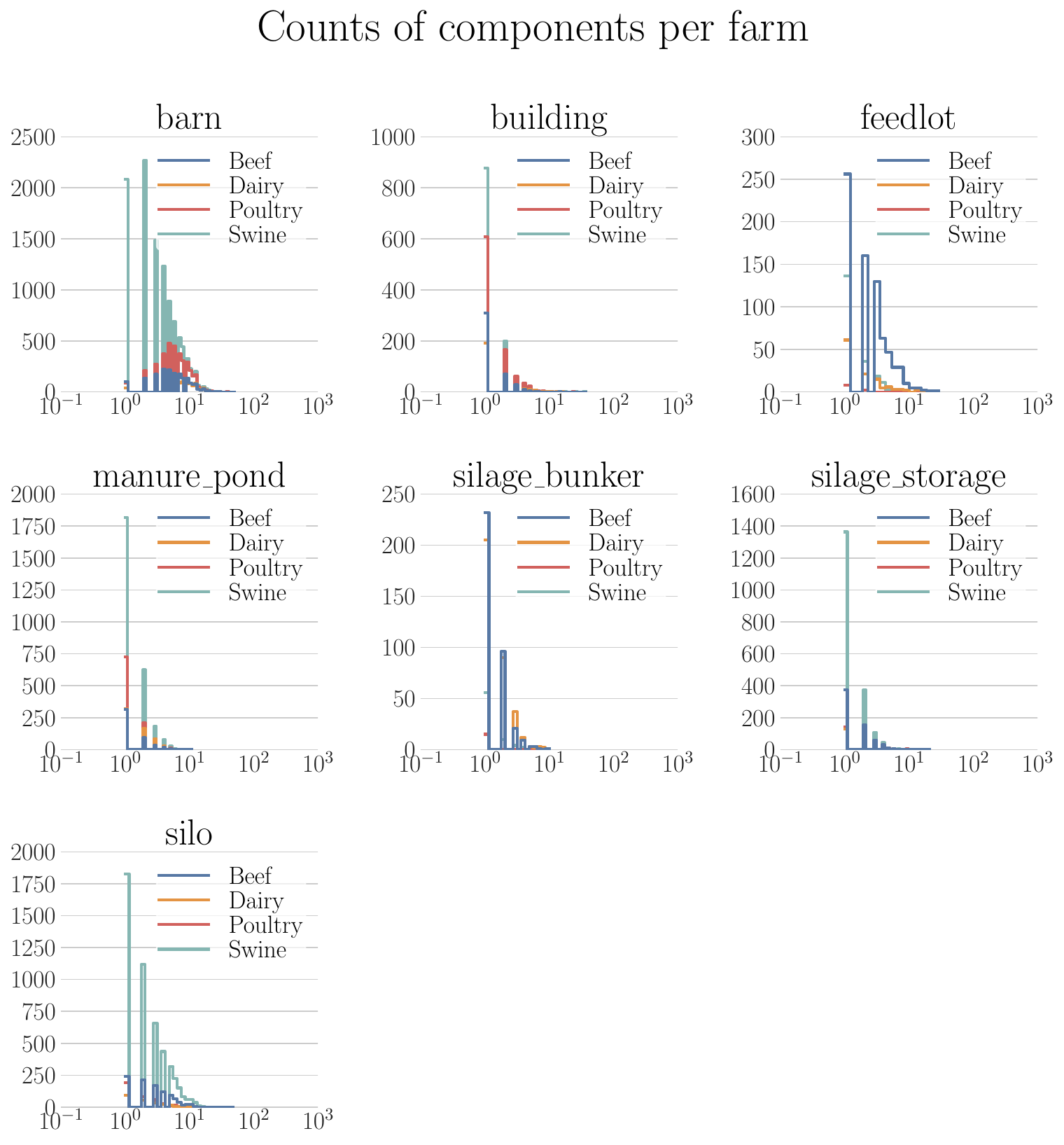}
    \caption{Statistics of components, and their sizes and shapes across
    livestock types. (i)~Area refers to the area of the bounding boxes
    across all positive patches that are identified as candidate
    infrastructure. (ii)~Aspect ratio refers to the ratio of the length of 
    the longer side to the length of the shorter side of the 
    bounding boxes. (iii)~For each CAFO patch, counts of each infrastructure
    type was computed in the third
    length to the shorter length. Each plot has subplots disaggregated
    by infrastructure type.}
    \label{fig:components}
\end{figure*}

\begin{table*}[h]
\centering
\caption{Summary of Core CAFO Data Sources Used in Livestock Detection.}
\label{tab:cafo-sources}
\begin{tabular}{p{2.3cm}p{6.5cm}p{4.5cm}}
\toprule
\textbf{Source} & \textbf{Description} & \textbf{Use} \\
\midrule
NAIP Imagery (2023) ~\cite{usda_naip} & High-resolution aerial imagery from the USDA National Agriculture Imagery Program & Visual input for CAFO patch extraction and model training \\ \hline
CAFOMaps \cite{cafomaps} & Multi-state (i.e., Alabama, Arkansas, Florida, Georgia, Louisiana, Mississippi, North Carolina, South Carolina, and Texas) labeled dataset containing 6,604 CAFOs with animal type annotations (e.g., poultry, swine, beef, dairy), curated by IOWA researchers & Ground-truth labels for training and validation of CAFO classification models \\\hline
State-CAFO Inventory \cite{INcafo,IAcafo,MDcafo,MIcafo,MNcafo,NYcafo,DEcafo} & Six independent data sources from official state-level CAFO registries 
corresponding to Indiana, Iowa, Maryland, Michigan, Minnesota, New York, and Delaware, curated from permit records, inspections, and nutrient management plans & Ground truth labels for ML-ready dataset \\\hline
Land Use Masks (NLCD) \cite{nlcd}& National Land Cover Database (NLCD) used for masking agricultural zones & Used to create negative samples \\
\bottomrule
\end{tabular}
\end{table*}
\end{document}